          [2001/04/25 1.1 (PWD)]
\documentclass[a4paper,twocolumn]{esapub2005} % European paper
\usepackage{times}
\usepackage[numbers]{natbib}
\usepackage{algpseudocode}
\usepackage{mathtools}
\usepackage{amsfonts}
\usepackage{comment}
\usepackage{amssymb}
\usepackage{amsmath} 
\usepackage{bm}
\usepackage{graphicx}
\usepackage{subcaption}

\title{Combining Parameter Identification and Trajectory Optimization: Real-time Planning for Information Gain}
\author{Keenan Albee$^{1*}$}
\author{Monica Ekal$^{2*}$}
\affil{Department of Aeronautics and Astronautics,
        Massachusetts Institute of Technology, 77 Massachusetts Avenue, Cambridge, MA 02139, USA; {\tt\small\{albee, linaresr\} at mit.edu}}
\author{Rodrigo Ventura$^2$}
\author{Richard Linares$^1$%
\thanks{The final version of this paper appears in the proceedings of the 15th Symposium on Advanced Space Technologies in Robotics and Automation (ASTRA) 2019. \newline \indent This work was completed under NASA Space Technology Research Fellowship support, grant number 80NSSC17K0077. This work was also supported by FCT project [UID/EEA/50009/2019], P2020 INFANTE 10/SI/2016, and an MIT Seed Project under the MIT Portugal Program. The authors gratefully acknowledge these sponsors.}% <-this % stops a space
\thanks{*Both authors contributed equally to this work.}% <-this % stops a space}
}
\affil{Institute for Systems and Robotics, Instituto Superior T\'ecnico, Av. Rovisco Pais 1, Lisboa 1049-001, Portugal; {\tt\small\{mekal, rodrigo.ventura\} at isr.tecnico.ulisboa.pt}}

\begin{document}


\maketitle

\begin{abstract}
Robotic systems often operate with uncertainties in their dynamics, for example, unknown inertial properties. Broadly, there are two approaches for controlling uncertain systems: design robust controllers in spite of uncertainty, or characterize a system before attempting to control it. This paper proposes a middle-ground approach, making trajectory progress while also accounting for gaining information about the system. More specifically, it combines excitation trajectories which are usually intended to optimize information gain for an estimator, with goal-driven trajectory optimization metrics. For this purpose, a measure of information gain is incorporated (using the Fisher Information Matrix) in a real-time planning framework to produce trajectories favorable for estimation. At the same time, the planner receives stable parameter updates from the estimator, enhancing the system model. An implementation of this learn-as-you-go approach utilizing an Unscented Kalman Filter (UKF) and Nonlinear Model Predictive Controller (NMPC) is demonstrated in simulation. Results for cases with and without information gain and online parameter updates in the system model are presented.
\end{abstract}

\begin{comment}
[longer abstract]
Robotic systems often operate with uncertainties in their dynamics, for example, unknown inertial properties. Broadly, there are two approaches for controlling these uncertain systems: operate under assumptions on the uncertain system and design robust controllers in spite of uncertainty, or characterize a system to fine-tune a controller before attempting system control. This paper proposes a middle-ground approach, making trajectory progress while also accounting for gaining information about a system, effectively blending these two paradigms. More specifically, it combines excitation trajectories which are usually intended to optimize information gain for an estimator, with goal-driven trajectory optimization metrics. For this purpose, a measure of information gain is incorporated (using the Fisher Information Matrix) in a real-time planning framework to produce trajectories favorable for a recursive estimator. At the same time, the planner receives stable parameter updates from the estimator, enhancing the system model. An implementation of this learn-as-you-go approach utilizing an Unscented Kalman Filter (UKF) and Nonlinear Model Predictive Control (NMPC) is demonstrated in simulation. Future experiments with the complete approach are laid out for potential microgravity testing on the SPHERES platform and simulation scenarios for robotic space applications.
\end{comment}

\section{Introduction}
Complete characterization of a system is often needed to conduct precise tasks like maneuvering in cluttered or sensitive environments, or performing close proximity operations such as docking. Such tasks are especially important for robotic spacecraft operations, where executing trajectories accurately is important, as is characterizing an unknown system, for example, after grappling unknown targets. In the literature, parameter identification methods used for spacecraft involve offline calculation of excitation trajectories that make the parameters observable \cite{shin1994}, \cite{christidi2017}, \cite{ekal2018inertial}. However, a pre-computed trajectory might not be feasible for robots operating in dynamic environments, like a free-flying robot transporting payload onboard a manned space station. It can also be undesirable to halt the system's operation for the identification process to be performed. Further, there are potential performance improvements from having better parameter estimates available, faster.

\begin{figure}[bt!]
    \centering
    \includegraphics{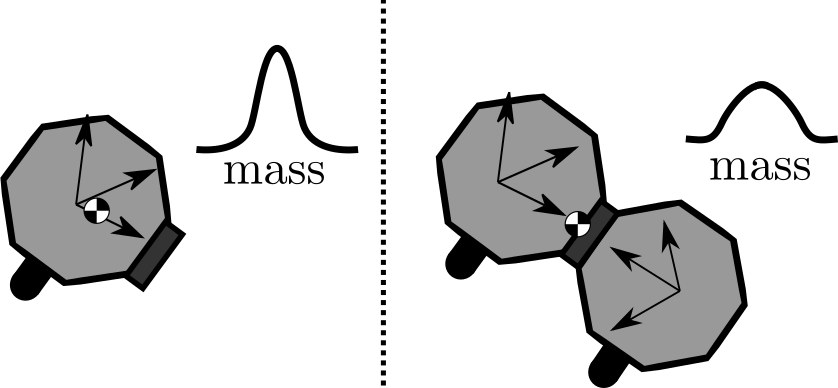}
    \caption{Robotic systems often have a variety of uncertain parameters. In particular, space systems that must manipulate or dock with an uncertain target (left) will inherit the uncertainty of the target (right) as the system's inertial properties change (e.g. mass, shown here with a sample probability density function).}
    \label{fig:my_label}
\end{figure}

To that end, this paper presents a framework for joint operation of the estimator and the planner, such that excitation trajectories (``richness") are combined with trajectory optimization in a real-time fashion \cite{Rao1998}, \cite{wilson2015real}, resulting in a weighting of information gain when parameter estimation is desired. Due to this real-time capability, the estimates can be incorporated by the planner on-the-fly, resulting in more precise execution. When sufficient confidence in the estimates is attained, the fast receding horizon planner allows the excitation trajectory to gradually blend into a quadratic-weighted optimal trajectory.

 This approach is well-suited in the context of certain space robotics applications like free-flying grappling of unknown targets. The novelty of this approach is that the robot's primary tasks can be performed concurrently with the identification process. In most parameter identification methods proposed in the literature, not only are excitation trajectories planned offline, but calibration is carried out exclusively--updating the controller model with the new estimates is not addressed. Other planning approaches (e.g. robust control) assume bounded uncertainty and provide guarantees within this uncertainty bound, but they do not address parameter estimation. \cite{Majumdar2017} \cite{Singh2017} Adaptive control does self-tune or adapt parameters online, but these methods track a reference trajectory or operating point rather than performing real-time planning and require trajectory ``richness" for adaptation. \cite{Slotinea} \cite{Wang2014}

 The paper is structured as follows: Section \ref{Prob Form} explains the fundamental problem, while Section \ref{Approach} presents details regarding the proposed solution, like the planning algorithm and information weighting. Section \ref{Implementation} elaborates on the simulation conditions under which the proposed algorithm was tested. The results of these simulation tests are provided in Section \ref{Results}. Concluding remarks and possible extensions of this work are provided in Section \ref{Conclusion}.

\section{Problem Formulation} \label{Prob Form}

A robotic system with state $\mathbf{x} \in \mathbb{R}^{n \times1}$ and uncertain parameters $\bm{\theta} \in \mathbb{R}^{j\times1}$ is initially positioned at state $\mathbf{x}_0$. A goal region $\mathcal{X}_g$ is specified (green dot in Figure 3). (For a rigid body system, $\bm{\theta}$ may include: mass, moments and products of inertia, and center of mass.) The dynamics and measurement model of the system are specified (equations (1) and (2), respectively). The aim is to plan a trajectory respecting constraints while minimizing a cost function $J$, formally stated in equation (11), by selecting optimal inputs to the dynamic system, $u_i \in \mathbf{u} \in \mathbb{R}^{k\times1}$. \cite{Paden2016}

\begin{figure}
    \centering
    \includegraphics[width=1.0\linewidth]{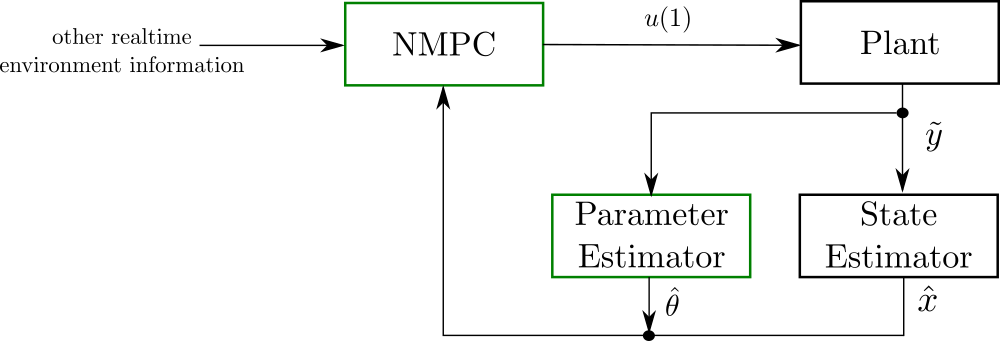}
    \caption{A high-level representation of the interaction between the NMPC and the estimation components. Components of interest are highlighted in green.}
    \label{fig:high-level rep}
\end{figure}

The problem of combining information gain with goal-driven planning by considering a robot with uncertain model parameters (e.g. a docked satellite) must be solved by: incorporating a measure of information gain; adapting parameters; adjusting information gain weighting; and performing real-time computation of $u_i^*$. At a high level, these components interact as shown in Figure \ref{fig:high-level rep}. The aim is to not only control the satellite to reach the goal position despite this uncertainty, but also to learn the system parameters while tracking a path towards the goal. Common sensor fusion algorithms combine measurements from different sensors to produce estimates of $\mathbf{x}$. It is assumed that these estimates are available for use by the controller and the parameter estimator.

\section{Approach} \label{Approach}

This work proposes combining the excitation that is obtained via offline excitation trajectories with traditional goal-oriented model predictive control. The result is a real-time planning framework that adds excitation to goal-achieving trajectories in order to characterize the system to satisfaction during useful motion. The Fisher information matrix is used as a measure of the information content, and a receding horizon planner is used for real-time planning in combination with a parameter estimator. Three key advantages are obtained:
\begin{itemize}
    \item Inclusion of excitation in goal-achieving trajectories
    \item Update of (more accurate) parameters on-the-fly
    \item Continual replanning for trajectory deviation
\end{itemize}
Figures 4 and 5 detail each of these advantages. This section provides an overview of the planning algorithm and its components for a general system.

\subsection{The Fisher Information Matrix}
Fisher information is a measure of the amount of information provided by the measurements on the unknown parameters. The Cram\'er-Rao Lower Bound (CRLB) gives a  theoretical lower bound on the expected  uncertainty of the parameter estimates. For an unbiased estimator, the CRLB is the inverse of the Fisher information matrix (FIM) \cite{crassidis2004optimal}.  If the information content is maximized, for instance, by maximizing any norm of the FIM, then it would also correspond to minimizing the lower bound on the variance of the estimates. Wilson et al. address using the FIM for excitation in \cite{Wilson2014}. The FIM is additive across time steps, and is at least positive semi-definite. The FIM has been used to maximize information in applications such as target localization \cite{ponda2009trajectory} and recently real-time active parameter estimation \cite{wilson2015real}.

\subsection{Calculation of the FIM}
Let the process and measurement models of the system be represented as:
\begin{align}
       \dot{\mathbf{x}} = f(\mathbf{x},\mathbf{u},\pmb{\theta}) + \mathbf{w}_x\\
       \Tilde{\mathbf{y}} = h(\mathbf{x},\mathbf{u},\pmb{\theta}) + \mathbf{w}_y
\end{align}
where the state vector is $\mathbf{x}$, the vector of the measured quantities is $\Tilde{\mathbf{y}} \in \mathbb{R}^{m \times1}$, and $\pmb{\theta} $ is the vector that contains the parameters to be estimated, with covariance $\mathbf{P} \in \mathbb{R}^{j \times j}$. The process noise $\mathbf{w}_x$, and measurement noise $\mathbf{w}_y$, are assumed zero mean Gaussian, with covariances $\mathbf{C} \in \mathbb{R}^{n \times n}$ and $\mathbf{\Sigma} \in \mathbb{R}^{m \times m}$, respectively. 
By definition, the FIM can be found as:

\begin{equation} \label{eq:FIM}
    F = E \left\{ \left[\frac{\partial}{\partial\pmb{\theta}}\ln{[p(\Tilde{\mathbf{y}}|\pmb{\theta}})] \right] \left[\frac{\partial}{\partial\pmb{\theta}}\ln{[p(\Tilde{\mathbf{y}}|\pmb{\theta}})] \right] ^T\right\}
\end{equation}
Assuming that there is no process noise in the parameter model, i.e., $\pmb{\theta}(t_i+1) = \pmb{\theta}(t_i)$, and due to the Gaussian nature of the measurement noise, over time $t_0$ to $t_{f}$ (3) reduces to:
\begin{equation}
    \mathbf{F} = \sum_{i = 0} ^{f}\mathbf{H}(t_i)^T \mathbf{\Sigma}^{-1} \mathbf{H}(t_i)
\end{equation}
where {\footnotesize \begin{align}
\mathbf{H}(t_i) = \frac{\partial h(\mathbf{x}(t_i),\mathbf{u}(t_i),\pmb{\theta}) }{\partial \pmb{\theta}} + \frac{\partial h(\mathbf{x}(t_i),\mathbf{u}(t_i),\pmb{\theta}) }{\partial \mathbf{x}}\cdot\frac{\partial \mathbf{x}(\mathbf{x}(t_i),\mathbf{u}(t_i),\pmb{\theta}) }{\partial \pmb{\theta}}
\end{align}  }
Let \begin{equation} \label{dx/dtheta}
   \pmb{ \phi}(t_i)= \frac{\partial \mathbf{x}(\mathbf{x}(t_i),\mathbf{u}(t_i),\pmb{\theta}) }{\partial \pmb{\theta}}
\end{equation} 
Eq. (\ref{dx/dtheta}) can also be written as (note that subscript $t_i$ is dropped for brevity):
\begin{equation}
\pmb{ \phi}=
\left[ \begin{array}{ccc}
\frac{\partial x_1}{\partial \theta_1} & \cdots & \frac{\partial x_1}{\partial \theta_j}\\
\vdots & \ddots & \vdots \\
\frac{\partial x_n}{\partial \theta_1} & \cdots & \frac{\partial x_n}{\partial \theta_j}
   \end{array} \right ]
\end{equation} 
Then, the value of $\pmb{\phi}$ can be obtained by the solution of the following ordinary differential equation, with initial conditions $ \pmb{\phi}(0) = \{0\}\in \mathbb{R}^{n \times j} $ \cite{4046308} \cite{wilson2015real} 
\begin{equation}
 \small{\pmb{\dot{\phi}} = \frac{\partial f(\mathbf{x},\mathbf{u},\pmb{\pmb{\theta}})}{\partial \mathbf{x}}\pmb{\phi} + \frac{\partial f(\mathbf{x},\mathbf{u},\pmb{\theta}) }{\partial \pmb{\theta}} }
\end{equation}

\begin{comment}
This computation can be performed using a numerical differential equation solver propagating $\mathbf{x}$, by appending $\pmb{\phi}$ as an additional state:

\begin{equation}
\mathbf{x}' = \begin{bmatrix}\mathbf{x}^\top \pmb{\phi}^\top \end{bmatrix}^\top
\end{equation}

Note that $\pmb{\phi}$ must be reconfigured to fit in the column vector of states.
\end{comment}

\subsection{Trajectory Optimization and Model Predictive Control}
A large body of work uses model predictive control for spacecraft operations, e.g. \cite{Weiss2015} \cite{Park2011}. Receding (or moving) horizon controllers function on the principle of only executing the first $n$ of $N$ timesteps of computed trajectory inputs. At every $n$ timesteps, an optimal trajectory is recomputed over a horizon of length $N$. Only $\mathbf{u}(1) ... \mathbf{u}(n)$ of these optimal inputs are actually executed. Since the optimization is computed online in real-time as frequently as every timestep, $n$=1, new information about the model and the environment can be incorporated on-the-fly. Moreover, the online trajectory generation also serves as a controller, making the system robust to error and uncertainty. Guarantees on stability and robustness are possible to provide for model predictive control (MPC), which uses quadratic cost and linear state dynamics, but are difficult to obtain for nonlinear MPC (NMPC).

MPC is particularly suited for online control in the presence of constraints as long as the problem being solved is a quadratic program. Nonlinear dynamics and cost throw out these assumptions--NMPC must be used, which relies, ultimately, on nonlinear programming (NLP) to solve a nonlinear optimization description of the system trajectory.

\begin{figure}[h]
    \centering
    \includegraphics[width=1.0\linewidth]{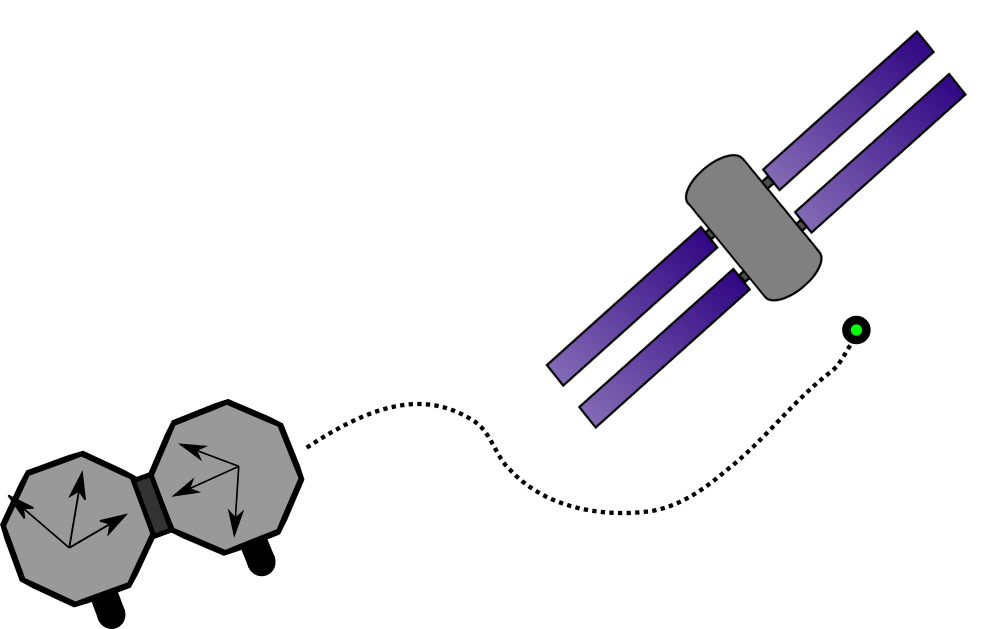}
    \caption{The fundamental trajectory optimization problem. A system, here two docked satellites, must obey dynamic constraints and environmental constraints (e.g. avoiding the space station shown here) while moving toward the goal, $\mathcal{X}_g$ (green dot).}
    \label{fig:my_label}
\end{figure}

\begin{figure*}[htb!]
    \centering
    \includegraphics[]{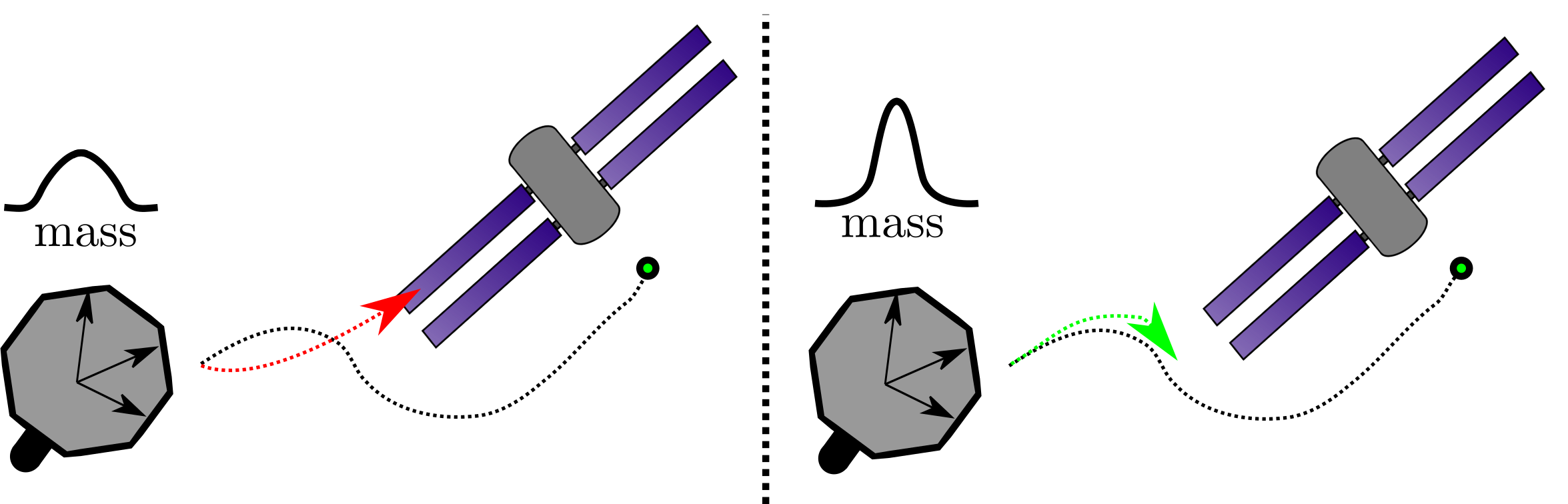}
    \caption{Real-time parameter updates allow for better decision-making since a more accurate model is available. The poorly known parameter impedes tracking (left), while online update improves tracking performance (right). The latest information from parameters and the environment can be taken in using this real-time implementation.}
    \label{fig:benefit}
\end{figure*}

\begin{figure*}[hbt!]
    \centering
    \includegraphics[]{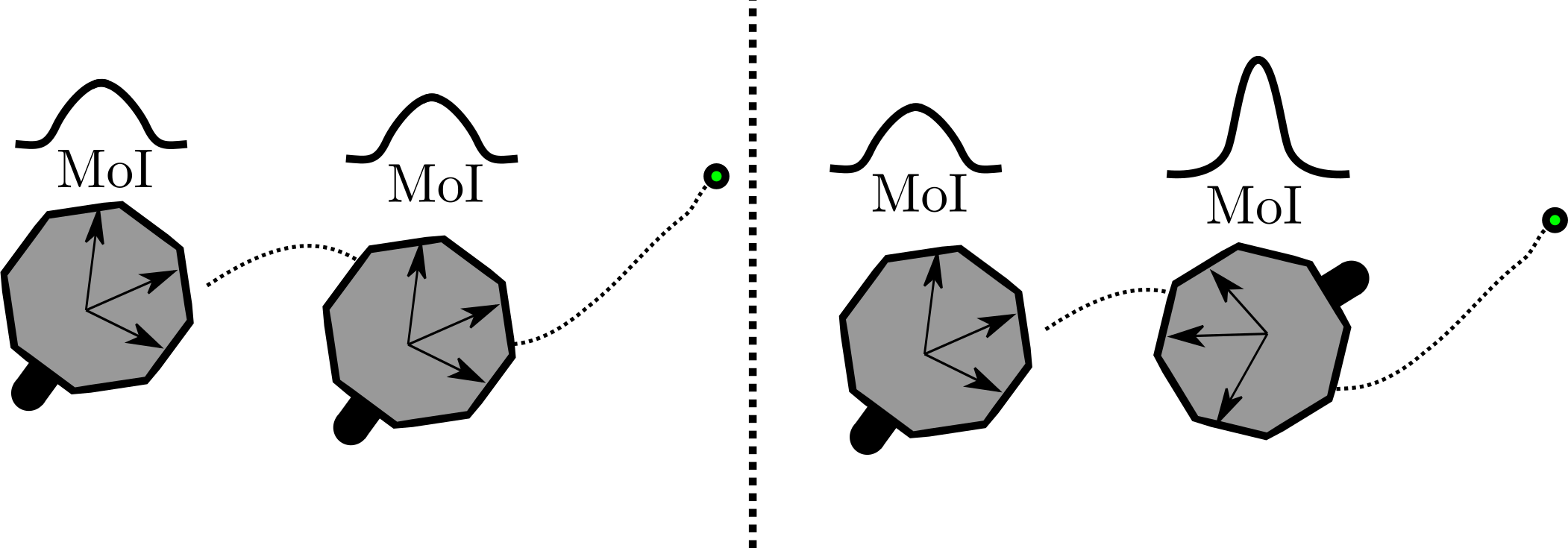}
    \caption{Excitation (information gain weighting) allows uncertain parameters to be better understood. The non-exciting trajectory has no need to apply torque so it gains no information about its moment of inertia (left). The addition of excitation allows tracking to the goal, but with additional rotation to understand moment of inertia (right).}
    \label{fig:benefit}
\end{figure*}

 \subsection{Cost Function and Optimization Formulation}\label{Cost Function}
The cost function combines two contrasting objectives of maximizing the excitation and minimizing the state error toward the goal. The FIM must be interpreted in scalar form to be incorporated in the excitation portion. Multiple ways of converting the information in the FIM to a scalar exist. \cite{faller2003simulation}. Minimizing the trace of the FIM inverse, i.e., $Tr\left(\mathbf{F}^{-1}\right)$, also referred to as the A-optimality criterion, is chosen in this case, equivalent to minimizing the lengths of the axes of the estimate uncertainty ellipsoid.

\begin{comment}
[FIM comments]
[The FIM is a symmetric and at least positive semi-definite, and usually scalar measures are used to incorporate Fisher information in a cost function.   "The  A-optimal  design  tries  to  maximize  the  trace  ofthe Fisher information matrixF. However, this criterion israrely used since it can lead to noninformative experiments
Faller, Klingmüller, and Timmer[16], with a covariance matrix that is not positive definite. A D-optimal design minimizes the determinant of the covariance matrix and can thus be interpreted as minimizing the geometric mean of the errors in the parameters. The largest error is minimized by the E-optimal design, which corresponds to a minimization of the largest eigenvalue of the covariance matrix. The modified E-optimal design minimizes the ratio of the largest to the smallest eigen vectorand thus optimizes the functional shape of the confidence intervals" ]
Trace... A optimality.
\\What does the trace of FIM mean? ( sum of eigen values of FIM, square of eigen value = length of each axis of the uncertainity ellipsoid.) ,WHy did we choose it?]
\end{comment}

The second part of the objective function is goal-driven, including tracking error and input with relative weighting matrics, $\mathbf{Q}$ and $\mathbf{R}$. The adjustable relative weighting term, $\gamma$, automatically begins to assign relative weight to this state-input weighting at a desired rate.

\begin{equation}
J = \int_{0}^{\infty}{\mathbf{x}^\top \mathbf{Q} \mathbf{x} + \mathbf{u}^\top \mathbf{R} \mathbf{u} + \gamma Tr\left(\mathbf{F}^{-1}\right)} \ dt
\end{equation}

Discretizing the above using the discrete version of the FIM, adding constraints, and setting $\mathbf{R}$ to 0 for this implementation gives:

\begin{equation}
\begin{aligned}
& \underset{\mathbf{u}}{\text{minimize J}} =
& & \sum_{i = 0}^{f}{\mathbf{x}^\top(t_i) Q \mathbf{x}(t_i) + \gamma Tr\left(\mathbf{\mathcal{\mathbf{F}}}(t_i)^{-1}\right)}\\
& \text{subject to}
& & \dot{\mathbf{x}} = f(\mathbf{x},\mathbf{u},\pmb{\theta}), \\
&&& \mathbf{u}_{min} \leq \mathbf{u}(t_i) \leq \mathbf{u}_{max}\\
&&& \mathbf{x} \in \mathcal{X}_{free}\\
\end{aligned}
\end{equation}

Note that  $f$ is the horizon length. In the implementation discussed here, $\gamma$ depends on the norm of state error as well as elapsed time through the following relation,

 \begin{equation}
     \gamma = e^{-1/\tau t} + \vert \vert \mathbf{x} \vert \vert_2
 \end{equation}
 where $\tau$ is the time constant for decay. As a result, the emphasis on the FIM weighting decreases exponentially, and is practically zero by the time the goal position is reached to avoid wasteful excitation. Developing a good heuristic for $\gamma$ is a matter of design choice: potentially useful choices include how much information gain is obtained compared to past values and trajectory-tracking error.

\section{Implementation} \label{Implementation}
As a test case, a free-flyer with uncertain mass and principal moments of inertia is considered. In such a situation, the benefits of using the proposed algorithm are two-fold: the information weighting will lead to an improvement in estimation accuracy, while updating these parameters in the NMPC system model should result in more accurate tracking. In order to highlight these aspects, the performance of this method was evaluated through two kinds of tests: 

\begin{enumerate}
    \item Comparison of the estimates when the cost function also weights information, as opposed to exclusively performing goal-tracking ($\gamma = 0$).
     \item Comparison of tracking performance with and without updating the system model with parameters estimated on-the-fly.
\end{enumerate}

\begin{figure}[!htb]
	\label{global}
	\begin{algorithmic}[1]
		\Procedure{Param-Plan}{$\mathbf{x}_0, \mathcal{X}_g, \mathbf{P}_0, \mathbf{Q}, \mathbf{R}, \bm{\theta}_0$}
		\State \texttt{InitEst($\mathbf{P}_0, \bm{\theta}_0)$}
		\State \texttt{InitNMPC($\mathbf{x}_{ref}, \mathbf{u}_{ref}$)}
		\While {$\mathbf{x}_k \not\in \mathcal{X}_g $}
			\State $u_{k+1} \gets \texttt{NmpcStep}(\mathbf{x}_k, \bm{\theta}_k, \mathbf{x}_{ref}, \mathbf{u}_{ref})$
		    \State $\mathbf{\dot{x}} = f(\mathbf{x}_k, \mathbf{u}_{k+1})$  \Comment{system dynamics update}
		    \State ($\bm{\theta}_{k+1}, \mathbf{P}_{k+1}) \gets \texttt{ParamEst} (\bm{\theta}_k, \mathbf{P}_k, \mathbf{\tilde{y}}_{k+1}, \mathbf{u}_k)$
		\EndWhile
		\State \Return {$(\mathcal{X}_{0:k}, \mathcal{U}_{0:k})$}
		\EndProcedure
		\State
		
		\Procedure{\texttt{NmpcStep}}{$\mathbf{x}_{k}, \bm{\theta}_k, \mathbf{x}_{ref}, \mathbf{u}_{ref}$}
		    \State $\gamma \gets \texttt{SetGamma}$
		    \State $u_{k+1} = \texttt{RunNmpc(CalcFisher)}$
		    \State \Return{$u_{k+1}$}
		\EndProcedure
	\end{algorithmic}
	\caption{An abbreviated summary of the planner's logic. The estimator and planner are initialized, and a desired target set $\mathcal{X}_g$ is set. At each step, a trajectory optimization is computed online with the latest $\bm{\theta}$. The $\gamma$ modification logic can be modified to desired operation in $\texttt{SetGamma}$.}
\end{figure}
 \begin{figure*}[hbt!]
    \centering
    \includegraphics[width=1.0\linewidth]{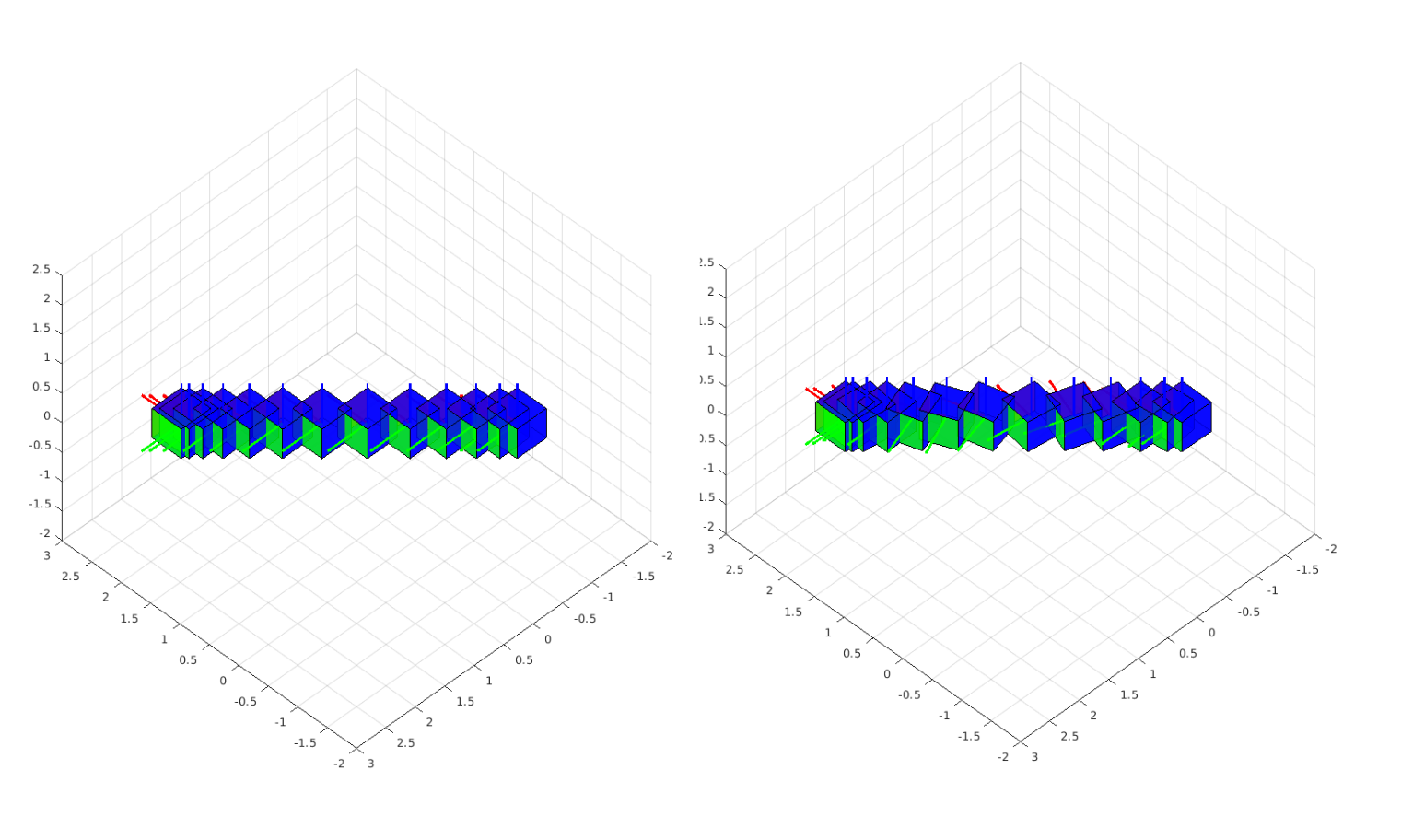}
    \caption{Two different trajectories for a system with uncertainty in mass and moment of inertia, $I_{zz}$. On the left, the non-excited trajectory gains no information about $I_{zz}$. However, on the right torques are commanded to aid the parameter estimator when information weighting is added. Note that the motion is 3DoF in this example for clarity.}
    \label{fig:labeled-wiggle}
\end{figure*}

\begin{figure*}[hbt!]
    \centering
    % \begin{subfigure}[b]{0.33\textwidth}
    %     \includegraphics[width=\textwidth]{Ixx.jpg}
    %     \caption{}
    % \end{subfigure}
    %  \begin{subfigure}[b]{0.33\textwidth}
    %     \includegraphics[width=\textwidth]{Iyy.jpg}
    %     \caption{}
    % \end{subfigure}
    %   \begin{subfigure}[b]{0.33\textwidth}
    %     \includegraphics[width=\textwidth]{Izz.jpg}
    %     \caption{}
    % \end{subfigure}
    \makebox[\textwidth][c]{\includegraphics[width=1.3\textwidth]{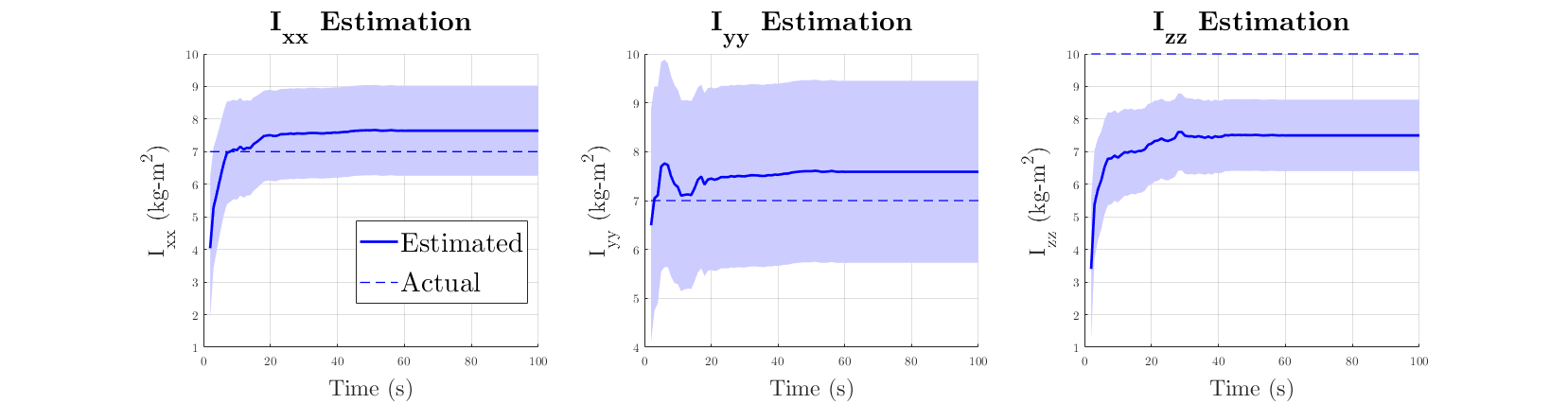}}
    \caption{Evolution of the estimates of the principal moments of inertia of a 6DoF free-flyer when the cost function does not have information weighting, i.e. $\gamma = 0$.}\label{fig:estimation_no_excitation}
\end{figure*}

These tests were carried out using a six degree of freedom (DoF) free-flyer model. The optimization problem detailed in Section \ref{Cost Function} was solved for each horizon using the ACADO toolkit \cite{Houska2011a}. Since a sequential estimator which could accommodate systems with complex dynamic models just as easily as 6DoF rigid body dynamics is desired, the Unscented Kalman Filter (UKF) \cite{julier2004} is used for mass and principal moments of inertia estimation. As stated previously, it is assumed that the state estimates are available through independent sensor fusion algorithms, and the resulting linear and angular velocities are fed to the UKF as measured data.

\subsection{Robot Dynamics and Specifications}
The state vector for a free-flying rigid body is given in equation (\ref{state_space}), consisting of rigid body position $\mathbf{r}$, orientation $\mathbf{q}$, linear velocity $\mathbf{v}$, and angular velocity $\pmb{\omega}$.

\begin{equation} \label{state_space}
\begin{aligned}
\begin{split}
&\mathbf{r} &= 
\begin{bmatrix}r_x&r_y&r_z\end{bmatrix}^\top\\
&\mathbf{q} &=  
\begin{bmatrix}q_x&q_y&q_z&q_\theta\end{bmatrix}^\top\\
&\mathbf{v} &=
\begin{bmatrix}v_x&v_y&v_z\end{bmatrix}^\top\\
&\bm{\omega} &=
\begin{bmatrix}\omega_x&\omega_y&\omega_z\end{bmatrix}^\top\\
\end{split}
\quad \quad \mathbf{x} = 
\begin{bmatrix}\mathbf{r}\\\mathbf{q}\\\mathbf{v}\\\bm{\omega}
\end{bmatrix}
\end{aligned}
\end{equation}

\begin{gather}
\mathbf{\dot{r}}_{CoM} = \mathbf{v}\\
\mathbf{\dot{v}}_{CoM} = \frac{\mathbf{F}}{m}\\
\bm{\dot{\omega}} = -\mathbf{I}^{-1}\bm{\omega}\times\mathbf{I}\bm{\omega} + \mathbf{I}^{-1}\tau\\
^{I}_{B}\mathbf{\dot{q}} = \frac{1}{2} \bar{H}(^{I}_{B}\mathbf{q})^{\top} {^{B}\bm{\omega}_{IB}}\\
\end{gather}

Where $\mathbf{I} \in \mathbb{R}^{3\times3}$ is the second moment of inertia with respect to the center of mass, expressed in $\mathcal{F}_B$. Note that the quaternion convention is scalar last and $^{I}_{B}\mathbf{q}$ represents the attitude of the body frame with respect to the inertial frame.$^{B}\bm{\omega}_{IB}$ indicates $\mathcal{F}_B$ with respect to $\mathcal{F}_I$, expressed in $\mathcal{F}_B$.

In the simulated robot model, the actual system mass is taken as $9.7$ $kg$, while the principal moments of inertia $I_{xx}$ and $I_{yy}$  are taken as $7$ $kg\cdot m^2$,  and $I_{zz}$ as $10$ $kg\cdot m^2$. The NMPC loop was run at 1 $Hz$, with a horizon length of 40 $s$.

\section{Results} \label{Results}
Results from the tests detailed in Section \ref{Implementation} are presented here. To demonstrate visually, Figure \ref{fig:labeled-wiggle} illustrates the excitation introduced in the trajectory due to the presence of information weighting. This 3DoF example shows the robot moving linearly to the goal point (the origin) if no rotation error is present and no information weighting is used. However, when information weighting for parameters of mass and $I_{zz}$ is used, torques are commanded about the $Z$ axis as the robot progresses toward the goal. With the aid of these rotations, data to estimate the value of $I_{zz}$ can now be gathered.

\begin{figure*}[hbt!]
    \centering
     \makebox[\textwidth][c]{\includegraphics[width=1.3\linewidth]{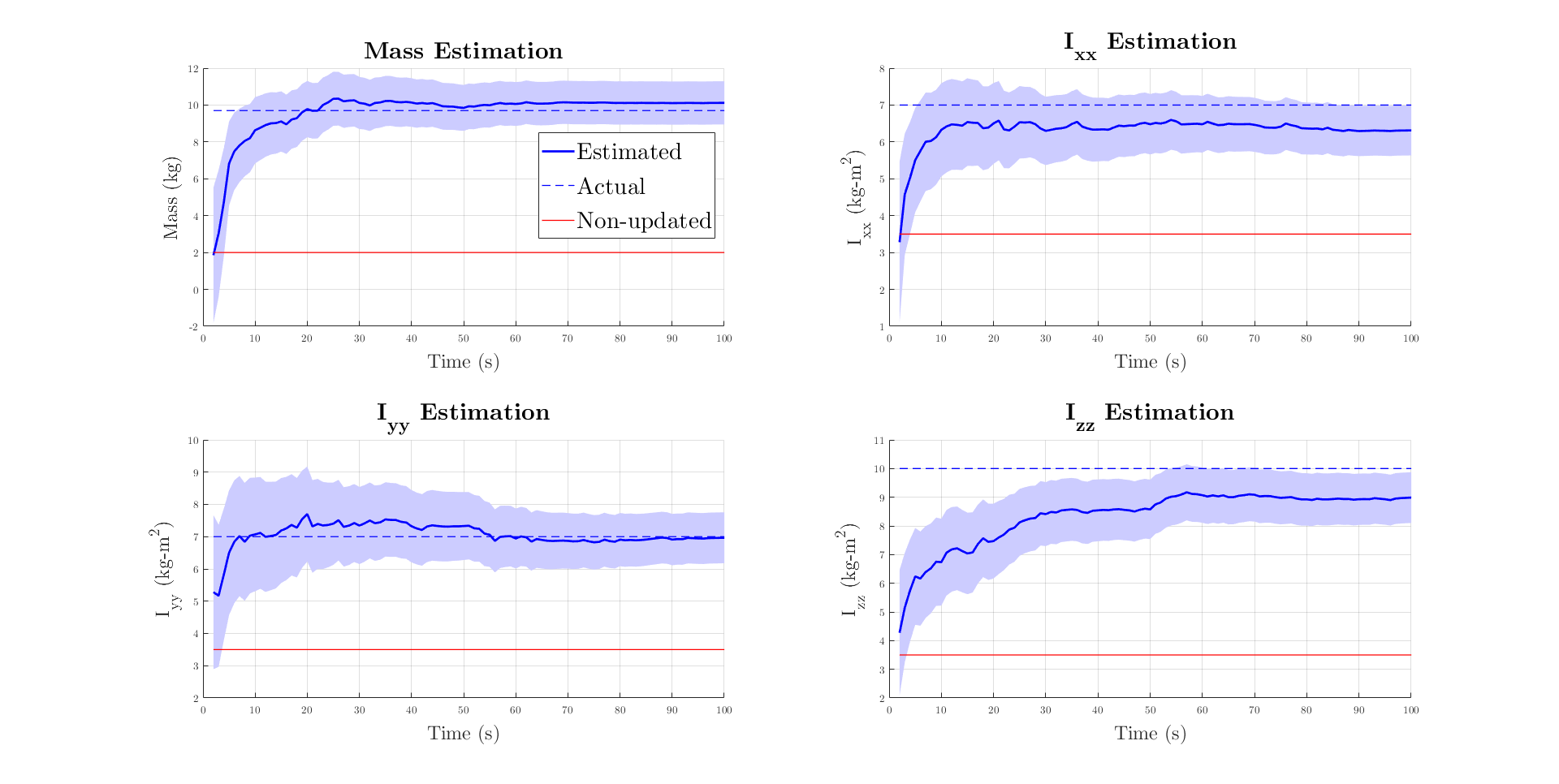}}
    \caption{Values of the inertial parameter (mass and principal moments of inertia) with respect to time, for the updated and non-updated case.}
    \label{fig:Res1_1_params}
\end{figure*}

For the first test objective, Figure \ref{fig:estimation_no_excitation} shows how the estimates of the principal moments of inertia of a 6DoF system evolve for a case of trajectory tracking without the inclusion of information weighting ($\gamma = 0$). From its initial state, the robot has to perform both rotation and translation motion to reach the goal. Even though some information about the moments of inertia is gained through these maneuvers, the resulting estimate is either less accurate, or is associated with a high level of uncertainty. In contrast, the 'Estimated' case in Figure \ref{fig:Res1_1_params} presents inertia estimates that are more accurate and certain when the cost includes information weighting, for the same start and goal points.

\begin{figure}[hbt!]
    \centering
    \includegraphics[width=1.0\linewidth]{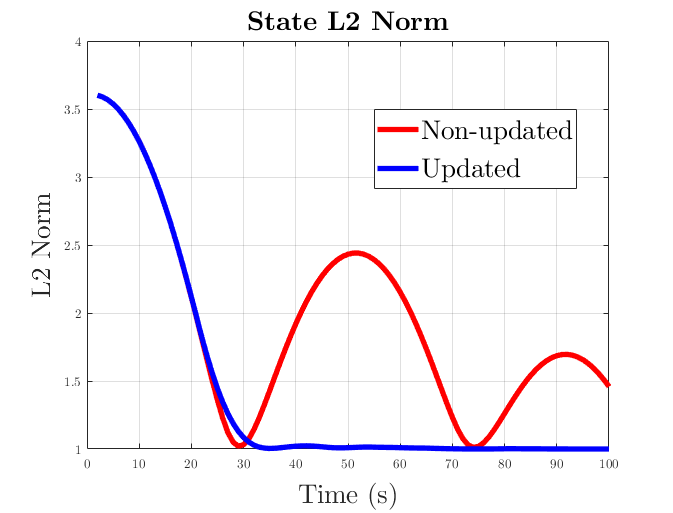}
    \caption{Trajectory tracking, with and without updates for the system used in Figure 9. With poor initial guesses for parameters and no updates, the system does not converge well to its desired attitude and position.}
    \label{fig:Res1_2_norm}
\end{figure}
\begin{figure}[hbt!]
    \centering
    \includegraphics[width=1.0\linewidth]{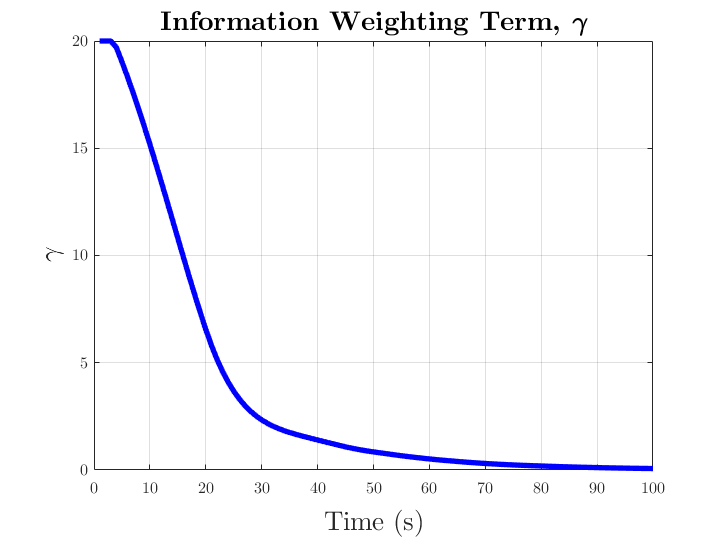}
    \caption{The information weighting term, $\gamma$, is decreased exponentially with time and inversely to distance to the goal state. Many intersting heuristics for $\gamma$ may be created, depending on the desired operation. The plot for test objective 2 is shown here.}
    \label{fig:gamma}
\end{figure}

 For the second test, i.e, assessment of NMPC performance with parameter updates,  the complete 6DoF model of the robot is considered. From an arbitrary position and attitude, the robot is commanded to reach the goal point where the attitude is the identity quaternion, $\mathbf{q} = [0, 0, 0, 1]^\top$ and all other states are zero. Figures \ref{fig:Res1_1_params} and \ref{fig:Res1_2_norm} illustrate that precise tracking is obtained when the known parameters are incorporated in the system model on-the-fly. In the case where the NMPC model parameters are not updated, the system does not stabilize at the goal even after 100 $s$. The simulation was performed on a 16 GB, 8 core Intel i7-4700MQ CPU @ 2.40GHz, running Linux Mint 18.3 using the C++ bindings of the ACADO MATLAB interface. NMPC steps are essentially instantaneous, requiring approximately 85 $ms$ on average for the stated 40 $s$ time horizon.

 \begin{comment}
 NMPC timing is on a personal laptop and will vary by scenario, but this captures roughly the speeds we see for this horizon in 6DoF (MATLAB tic toc)
 \end{comment}

\begin{comment}
In this section, results from simulations are presented. A scenario of robot calibration was presented. A (minimum-time/input?) Model Predictive Controller (MPC) was initialized with previously known values of the robot mass. During subsequent time steps, the mass model in the MPC was updated with latest estimates from the Unscented Kalman Filter (UKF) which performes joint state and parameter estimation (In this case, positions, velocities and robot mass). (...The plots show that the MPC tracking gets better as the updated value of the mass approaches the real one...)
https://www.overleaf.com/project/5c6467c6b5af075aa8e0ff93
\end{comment}

\section{Conclusion} \label{Conclusion}
In this paper, we introduce a framework for robotic systems to plan an optimal goal-oriented trajectory while simultaneously attempting to identify some of their uncertain parameters. The latest estimates, as well as information regarding the estimator's confidence in them, are incorporated in the real-time calculation of the trajectories. This enables the system to perform calibration while also using this information to plan for its primary task in a more robust manner. 

A cost function composed of the trace of the inverse Fisher Information Matrix (FIM) as a measure of information gain and the error of the current state to the goal was formulated. A Nonlinear Model Predictive Controller (NMPC) was used to minimize this cost, obeying system dynamics and actuator constraints. Its real-time nature (~85 $ms$ per 40 $s$ time horizon) allows for model updates based on new information from the estimator. Simulation results for 3DoF planar as well as complete 6DoF rigid body dynamics with uncertain inertial parameters were presented. Trajectories with little overshoot from arbitrary initial conditions were shown, particularly in comparison to planning that does not update parameters on-the-fly. In addition, trajectories for another scenario show the ability of information-weighted planning to discover parameters that would otherwise not be adequately excited by a standard cost function.

Development of experiments using the SPHERES testing platform onboard the International Space Station are under consideration, which would provide 6DoF hardware validation of the approach. In future work, this approach can be extended for systems with more complex models, for instance, control and characterization of docked satellites or manipulator-equipped satellites handling uncertain payloads. In cases where some parameters are known with more certainty than others, the proposed algorithm could be modified to re-plan trajectories by focusing only on the uncertain subset of parameters. Moreover, measures can be implemented to identify divergence of the parameter estimates or to monitor poor trajectory tracking. Developing useful heuristics for shifting the relative weighting term, $\gamma$ is another interesting area of future progress, along with a more thorough analysis of guarantees provided by this control method. Finally, these experiments have not yet demonstrated one of the key advantages of real-time trajectory optimization, obstacle avoidance. Additional useful test scenarios (e.g. with obstacles) would be worth analyzing.
\vspace{20mm}

\section*{Acknowledgments}

The authors would like to thank To\~{n}o Teran and William Sanchez for their help with experimentation using the SPHERES research testbed, along with Hailee Hettrick for editing assistance.

% The following bibliography was produced with
%   \bibliographystyle{aa}
%   \bibliography{esapub}
% The results are inserted directly here to simplify
% the demonstration.

\bibliographystyle{plain}
\bibliography{biblio}

\begin{thebibliography}{10}

\bibitem{christidi2017}
Olga-Orsalia Christidi-Loumpasefski, Kostas Nanos, and Evangelos Papadopoulos.
\newblock On parameter estimation of space manipulator systems using the
  angular momentum conservation.
\newblock In {\em 2017 IEEE International Conference on Robotics and Automation
  (ICRA)}, pages 5453--5458. IEEE, 2017.

\bibitem{crassidis2004optimal}
John~L Crassidis and John~L Junkins.
\newblock {\em Optimal estimation of dynamic systems}.
\newblock Chapman and Hall/CRC, 2004.

\bibitem{ekal2018inertial}
Monica Ekal and Rodrigo Ventura.
\newblock On inertial parameter estimation of a free-flying robot grasping an
  unknown object.
\newblock In {\em 2018 5th International Conference on Control, Decision and
  Information Technologies (CoDIT)}, pages 815--821. IEEE, 2018.

\bibitem{faller2003simulation}
Daniel Faller, Ursula Klingm{\"u}ller, and Jens Timmer.
\newblock Simulation methods for optimal experimental design in systems
  biology.
\newblock {\em Simulation}, 79(12):717--725, 2003.

\bibitem{Houska2011a}
B.~Houska, H.J. Ferreau, and M.~Diehl.
\newblock {ACADO} {T}oolkit -- {A}n {O}pen {S}ource {F}ramework for {A}utomatic
  {C}ontrol and {D}ynamic {O}ptimization.
\newblock {\em Optimal Control Applications and Methods}, 32(3):298--312, 2011.

\bibitem{Park2011}
{Hyeongjun Park}, Stefano {Di Cairano}, and Ilya Kolmanovsky.
\newblock {Model predictive control for spacecraft rendezvous and docking with
  a rotating/tumbling platform and for debris avoidance}.
\newblock {\em American Control Conference}, pages 1922--1927, 2014.

\bibitem{julier2004}
Simon~J Julier and Jeffrey~K Uhlmann.
\newblock Unscented filtering and nonlinear estimation.
\newblock {\em Proceedings of the IEEE}, 92(3):401--422, 2004.

\bibitem{Majumdar2017}
Anirudha Majumdar and Russ Tedrake.
\newblock {Funnel Libraries for Real-Time Robust Feedback Motion Planning}.
\newblock pages 1--60, 2017.

\bibitem{Paden2016}
Brian Paden, Michal Cap, Sze~Zheng Yong, Dmitry Yershov, and Emilio Frazzoli.
\newblock {A Survey of Motion Planning and Control Techniques for Self-driving
  Urban Vehicles}.
\newblock 1(1):33--55, 2016.

\bibitem{ponda2009trajectory}
Sameera Ponda, Richard Kolacinski, and Emilio Frazzoli.
\newblock Trajectory optimization for target localization using small unmanned
  aerial vehicles.
\newblock In {\em AIAA guidance, navigation, and control conference}, page
  6015, 2009.

\bibitem{Rao1998}
C.~V. Rao, S.~J. Wright, and J.~B. Rawlings.
\newblock {Application of interior-point methods to model predictive control}.
\newblock {\em Journal of Optimization Theory and Applications},
  99(3):723--757, 1998.

\bibitem{shin1994}
Jin-Ho Shin and Ju-Jang Lee.
\newblock Dynamic control with adaptive identification for free-flying space
  robots in joint space.
\newblock {\em Robotica}, 12(6):541--551, 1994.

\bibitem{Singh2017}
Sumeet Singh, Anirudha Majumdar, Jean~Jacques Slotine, and Marco Pavone.
\newblock {Robust online motion planning via contraction theory and convex
  optimization}.
\newblock {\em Proceedings - IEEE International Conference on Robotics and
  Automation}, pages 5883--5890, 2017.

\bibitem{Slotinea}
Jean-Jacques~E Slotine and Weiping Li.
\newblock {\em {Applied Nonlinear Control}}.
\newblock Prentice-Hall, Inc., Englewood Cliffs, NJ, 1991.

\bibitem{4046308}
J.~H. {Taylor}.
\newblock The cramer-rao estimation error lower bound computation for
  deterministic nonlinear systems.
\newblock In {\em 1978 IEEE Conference on Decision and Control including the
  17th Symposium on Adaptive Processes}, pages 1178--1181, Jan 1978.

\bibitem{Wang2014}
Hanlei Wang.
\newblock {Adaptive Control of Robot Manipulators With Uncertain Kinematics and
  Dynamics}.
\newblock 62(2):948--954, 2017.

\bibitem{Weiss2015}
Avishai Weiss, Morgan Baldwin, Richard~Scott Erwin, and Ilya Kolmanovsky.
\newblock {Model predictive control for spacecraft rendezvous and docking:
  Strategies for handling constraints and case studies}.
\newblock {\em IEEE Transactions on Control Systems Technology},
  23(4):1638--1647, 2015.

\bibitem{wilson2015real}
Andrew~D Wilson, Jarvis~A Schultz, Alex~R Ansari, and Todd~D Murphey.
\newblock Real-time trajectory synthesis for information maximization using
  sequential action control and least-squares estimation.
\newblock In {\em 2015 IEEE/RSJ International Conference on Intelligent Robots
  and Systems (IROS)}, pages 4935--4940. IEEE, 2015.

\bibitem{Wilson2014}
Andrew~D. Wilson, Jarvis~A. Schultz, and Todd~D. Murphey.
\newblock {Trajectory synthesis for fisher information maximization}.
\newblock {\em IEEE Transactions on Robotics}, 30(6):1358--1370, 2014.

\end{thebibliography}
\end{document}